\documentclass[review]{elsarticle}
\usepackage{booktabs}
\usepackage{multirow}
\usepackage{siunitx}
\usepackage{lineno,hyperref}
\modulolinenumbers[5]
\usepackage{amsmath}
\journal{Journal of \LaTeX\ Templates}

\begin{document}

\begin{frontmatter}

\title{Investigation and Analysis of Hyper and Hypo neuron pruning to selectively update neurons during Unsupervised Adaptation}

%% Group authors per affiliation:
\author{{Vikramjit Mitra$^1$\fnref{myfootnote1}}, {Horacio Franco}\fnref{myfootnote2}}
\address{$^1$ Dept. of Electrical and Computer Eng., University of Maryland, College Park, MD, USA}
\address{$^2$ Speech Technology and Research Laboratory, SRI International, Menlo Park, CA, USA}
\fntext[myfootnote1]{email: vmitra@umd.edu}
\fntext[myfootnote2]{email: horacio.franco@sri.com}
\fntext[myfootnote3]{This work was performed when the first author was at SRI International. He is currently affiliated with Apple.}
%% or include affiliations in footnotes:

\begin{abstract}
Unseen or out-of-domain data can seriously degrade the performance of a neural network model, indicating the model’s failure to generalize to unseen data. Neural net pruning can not only help to reduce a model’s size but can improve the model’s generalization capacity as well. Pruning approaches look for low-salient neurons that are less contributive to a model’s decision and hence can be removed from the model. This work investigates if pruning approaches are successful in detecting neurons that are either high-salient (mostly active or hyper) or low-salient (barely active or hypo), and whether removal of such neurons can help to improve the model’s generalization capacity. Traditional blind adaptation techniques update either the whole or a subset of layers, but have never explored selectively updating individual neurons across one or more layers. Focusing on the fully connected layers of a convolutional neural network (CNN), this work shows that it may be possible to selectively adapt certain neurons (consisting of the hyper and the hypo neurons) first, followed by a full-network fine tuning. Using the task of automatic speech recognition, this work demonstrates how the removal of hyper and hypo neurons from a model can improve the model’s performance on out-of-domain speech data and how selective neuron adaptation can ensure improved performance when compared to traditional blind model adaptation.
\end{abstract}

\begin{keyword}
\texttt{neural net pruning} \sep {unsupervised adaptation} \sep {convolutional neural network} 
\MSC[2010] 00-01\sep  99-00
\end{keyword}

\end{frontmatter}

%\linenumbers

\section{Introduction}

Deep Neural Networks (DNNs) have become the ubiquitous machine learning technique, demonstrating significant performance gains over its peers in almost all applications where they have been tested \cite{krizhevsky2012imagenet, he2016deep,mikolov2013efficient}, including automatic speech recognition (ASR) \cite{mohamed2011acoustic, seide2011conversational}. DNNs are both data hungry and data sensitive \cite{grezl2014further}.

Neural nets are usually overparameterized with significant redundancy in their number of neurons and the transforms that they learn, given training data \cite{louizos2017learning}. A consequence of information redundancy in neural nets is data over-fitting, where a model ends up learning fine grained structures present within the training data that may not be relevant for the classification task and may not be present in unseen data. Such over-fitting results in models failing to generalize well to unseen data conditions, as a consequence significant performance degradation is observed when such models are deployed to unknown datasets as compared to seen or in-domain datasets. To prevent models from over-fitting, well known techniques such as regularization, dropout \cite{srivastava2014dropout}, pruning \cite{hertz1991introduction}, data augmentation \cite{karafiat2015three} etc., are deployed. Pruning strategies try to minimize data over-fitting by removing less salient neurons from a model while retaining the model’s performance on a given data-set. The goal of network pruning is to find neurons in a model whose removal will not degrade the model’s performance, while improving its generalization capacity.

Using traditional pruning algorithms, this work investigates the role of neurons that are deemed as less-salient versus the ones that are high-salient. The work investigates the following:
\newline \textit{(1) Is the traditional approach of pruning out “less salient” neurons the best strategy for performing neural network pruning?
\newline (2) What are the effects of network pruning in each individual layer of a fully connected network? 
\newline (3) Is there any benefit in adapting only the “to-be-pruned” neurons at first during unsupervised model adaptation, followed by adapting the whole network?}

The general observation from this study is that there are three main groups of neurons: (1) \textit{less-salient}, or almost inactive or ‘\textit{hypo}’ active neurons; (2) \textit{high-salient}, or, very active or, ‘\textit{hyper}’ active neurons, and (3) \textit{medium-salient} neurons, or relevant neurons, whose removal results in catastrophic degradation in a network’s performance. .

Observations in this paper are based on a speech recognition task, with both ‘\textit{in-domain}’ or ‘\textit{seen}’ data and ‘\textit{out-of-domain}’ or ‘\textit{unseen}’ data as evaluation tasks. The in-domain evaluation data consists of speech recorded in close microphone conditions with varying microphone channels and background degradations. The out-of-domain evaluation data consists of speech recorded using distant microphones, recorded in varying room conditions with different reverberation times and noisy background conditions. 

We investigate a new mutual-information based neural net pruning approach, capable of detecting salient neurons leveraging temporal properties of a speech signal and demonstrate its effectiveness in a speech recognition task. Please note that proposing a new pruning approach is not the aim of this paper; instead, understanding which neurons can be pruned and whether there is any benefit in selectively updating such neurons during unsupervised model adaptation is the main goal of this work.

We derive our conclusions based on a state-of-the-art convolutional neural network (CNN) acoustic model \cite{mitra2015time}, focusing on network pruning only for the fully connected layers. Unlike prior work, we investigate layer-wise pruning where we evaluate the impact of pruning on each layer individually. 

We investigate if the neurons that are supposed to be pruned, can be used to selectively adapt the neural network model to out-of-domain data. This investigation stems from the fact that certain neurons can be pruned as they have been deemed as “not useful”, while the others are retained as they are useful. Rationally, it makes sense to retrain the ‘not useful’ neurons, while retaining the ‘useful’ neurons as-is. Our observations show that such selective model adaptation can result in well regularized models that can adapt to out-of-domain data while retaining their performance on in-domain data.

\section{Data}

The acoustic models in this work were trained using the multi-conditioned, noise and channel degraded training data from the 16 kHz Aurora-4 \cite{hirsch2002experimental} noisy WSJ0 corpus. Aurora-4 data contains six additive noise types with channel-matched and mismatched conditions. It was created from the standard 5K WSJ0 database, containing 15 hours of training data and 0.6 hours (330 utterances) of testing data. The test data is replicated into 14 test sets (0.6 hours each) consisting of two different channel conditions and six different added noises (car; babble; restaurant; street; airport; and train station) in addition to the clean condition. The signal-to-noise ratio (SNR) for the test sets varied between 0 and 15 dB. The evaluation set consists of 5K words. The Aurora-4 test set is used as the in-domain evaluation set in our experiments. 

In this work, we treated reverberation as the out-of-domain data condition (not included during training). For adaptation, optimization, and evaluation purposes, we have used the single-microphone subset of the training, dev, and the eval sets distributed with the REVERB 2014 (denoted as REVERB14 in this paper) \cite{kinoshita2013reverb} challenge, respectively. The adaptation set consists of the clean WSJCAM0 \cite{robinson1995wsjcamo} data, which was convolved with different room impulse responses (reverberation times from 0.1 to 0.8 sec) and then corrupted with background noise. The performance was evaluated on the dev and test subsets, which contain both real and simulated reverberation conditions. The real data is borrowed from the MC-WSJ-AV corpus \cite{lincoln2005multi}, which consists of utterances recorded in a noisy and reverberant room. The simulated and the real eval sets contained 1088 and 372 utterances, respectively, split equally between far and near microphone conditions. Note that none of our experiments used any speaker-level information.

\section{Acoustic Model}

Earlier \cite{mitra2014evaluating} we found that CNN models perform better than the DNNs for Aurora-4 ASR task, hence CNN acoustic models were used in this work. The training alignments consisted of 3125 context-dependent (CD) states, which were generated from a baseline DNN acoustic model. The input acoustic features were formed by using a context of 17 frames (8 frames on either side of the current frame). We observed that time-frequency convolution (using TFCNN \cite{mitra2015time,yilmaz2018articulatory,yilmaz2019articulatory, mitra2017speech,mitra2017hybrid,mitra2017robust}) performed better than 1-D frequency convolution, and hence we have focused on the TFCNN acoustic models for our experiments presented in this paper. The TFCNN architecture is same as in \cite{mitra2015time, mitra2016fusion}, where two parallel convolutional layers are used at the input, one performing convolution across time, and the other across frequency on the input filterbank features. The TFCNNs had 75 filters to perform time convolution and 200 filters to perform frequency convolution. For time and frequency convolution, eight bands were used, followed by a max-pooling over three samples after frequency convolution, and five samples for time convolution. The feature maps after both the convolution operations were concatenated and then fed to a fully connected DNN having 2048 nodes and four hidden layers. The TFCNN network was trained using four initial iterations with a constant learning rate of 0.008, followed by learning rate halving based on cross-validation error. Training stopped when no further reduction in cross-validation error was noted or when it started to increase. Backpropagation was performed by using stochastic gradient descent with a mini-batch size of 256 examples. Note that each model has been trained three times to interpret the uncertainty in the model performance. In our experiments we provide the mean WER from the multiple iterations and their standard deviations are reported within parenthesis in tables and as error bars in figures. Note that the baseline acoustic model was trained from a random initial seed, and that baseline model has been used as the seed model for all our experiments.

We used gammatone filter-bank energies (GFBs) as the acoustic feature for our experiments. GFBs were generated using a bank of 40 time-domain gammatone filters, using an analysis window of 26 millisecond and a frame rate of 10 millisecond. The gammatone subband powers were compressed using $15^{th}$ root.

\section{Neural Network Pruning}

One main goal of this work is to understand how layer-wise pruning affects performance on out-of-domain data and validate that across multiple pruning strategies. We explored three pruning strategies and they are briefly described below:
\leavevmode
\newline 
(1) \textit{Magnitude based pruning} (MBP): assumes that small weights are less important \cite{wan2009enhancing, hagiwara1994simple, sietsma1988neural}. Two factors are typically used for determining redundant neurons: consuming energy and weight power [20]. In this work, we ascertained a neuron’s saliency by using its weight’s power \cite{hagiwara1994simple}.
\leavevmode
\newline 
(2) \textit{Optimal Brain Surgeon} (OBS) is a sensitivity-based approach that attempts to find the contribution of each weight or node in the network and then prunes the weight or node that have the least effect on the objective function. One of the celebrated sensitivity-based pruning algorithms is Optimal Brain Damage (OBD) \cite{lecun1990optimal}, that approximates the measure of ‘saliency’ of a weight by estimating the second derivative of the network output error with respect to that weight. OBS \cite{hassibi1993optimal} relaxed some of the assumptions in OBD. In this work we have used OBS to ascertain neural saliency of each layer.
\leavevmode
\newline 
(3) \textit{Mutual Independence} (MI) based pruning leverage relationships between input and output of a neuron to predict the neuron’s saliency. One such approach is mutual information based \cite{xing2009two, zhang2010node}, which uses singular value decomposition to analyze the hidden unit activation covariance matrix, where the rank of the covariance matrix determines the optimal number of hidden units.

In this work we investigated a cross-correlation based MI measure to estimate the input-output relationship of a neuron to determine the saliency of that neuron, given an input feature set. The cross-correlation analysis can be performed using a subset of the training data after neural network training. 

Let ${a_{n,l}}$ be the nth neuron of the $l^{th}$ layer, whose inputs are:

${[x_{1,l-1}[t], x_{2,l-1}[t], ... x_{p,l-1}[t], ... x_{N,l-1}[t]]}$ 

where ${x_{p,l-1}[t]}$ is the activation from the $p^{th}$ neuron of the preceding ${l-1^{th}}$ layer at time instant $t$. Let the activation from the ${a_{n,l}}$ at $t$ be ${x_{n,l}[t]}$, assuming that ${l-1^{th}}$ layer had $N$ neurons, the mean absolute cumulative cross-correlation between the input and the output activations for the neuron ${a_{n,l}}$ over a time window $q$ at time $t$ will be:

\begin{equation}
\begin{aligned}
\hat{r}_{n,l}[t] = 1/N {\sum_{p=1}^{N}} \left| {\sum_{k=-(q/2-1)}^{q/2}{x_{n,l}[t]x_{p,l-1}[t_k]}} \right|
\end{aligned}
\label{eq:eq1}
\end{equation}

Saliency of neuron ${a_{n,l}}$ of layer $l$ is determined by the value of ${\hat{h}_{n,l}[t]}$; higher the value the more salient the neuron is and vice versa. Note that, this measure is particularly suitable for quasi-periodic pseudo-stationary signals like speech, that retain a certain degree of temporal correlation. The measure in (\ref{eq:eq1}) indicates that if the inputs to the neuron $n$ is not correlated to the output from the neuron $n$, then neuron $n$ is less salient, in the sense that is in not conveying information to the succeeding layers. ${\hat{r}_{(n,l)}[t]}$ is used to estimate saliency in the MI based pruning investigated in this paper.

Note that while pruning, we rank-sorted the neurons based on their saliency, where the top salient neurons are selected as high-salient or hyper-active neurons and the low-salient neurons are selected as low-salient or hypo-active neurons.

\section{Neural Network Pruning}
We investigated layer-wise pruning, focusing only on the fully connected hidden layers, by pruning 2\% to 12\% in steps of 2\%. We investigate layer-wise pruning in three possible ways: 
\newline (a) pruning neurons with low saliency, 
\newline (b) pruning neurons with high saliency, 
\newline (c) pruning neurons with medium saliency (i.e., the neurons that are in between (a) and (b). 

Note that during layer-wise pruning a 2\% pruning means pruning 2\% of the neurons in that layer, for example if the number of neurons in that layer is 2048, a 2\% pruning of that layer would indicate removing 41 neurons from that layer.

\subsection{Pruning of low salient (hypo) neurons}
\label{subsection1}
\begin{figure}[]
\begin{minipage}[b]{1.0\linewidth}
  \centering
  %\centerline{\includegraphics[width=8.5cm]{draft3_fig1}}
  \centerline{\includegraphics[width=10cm]{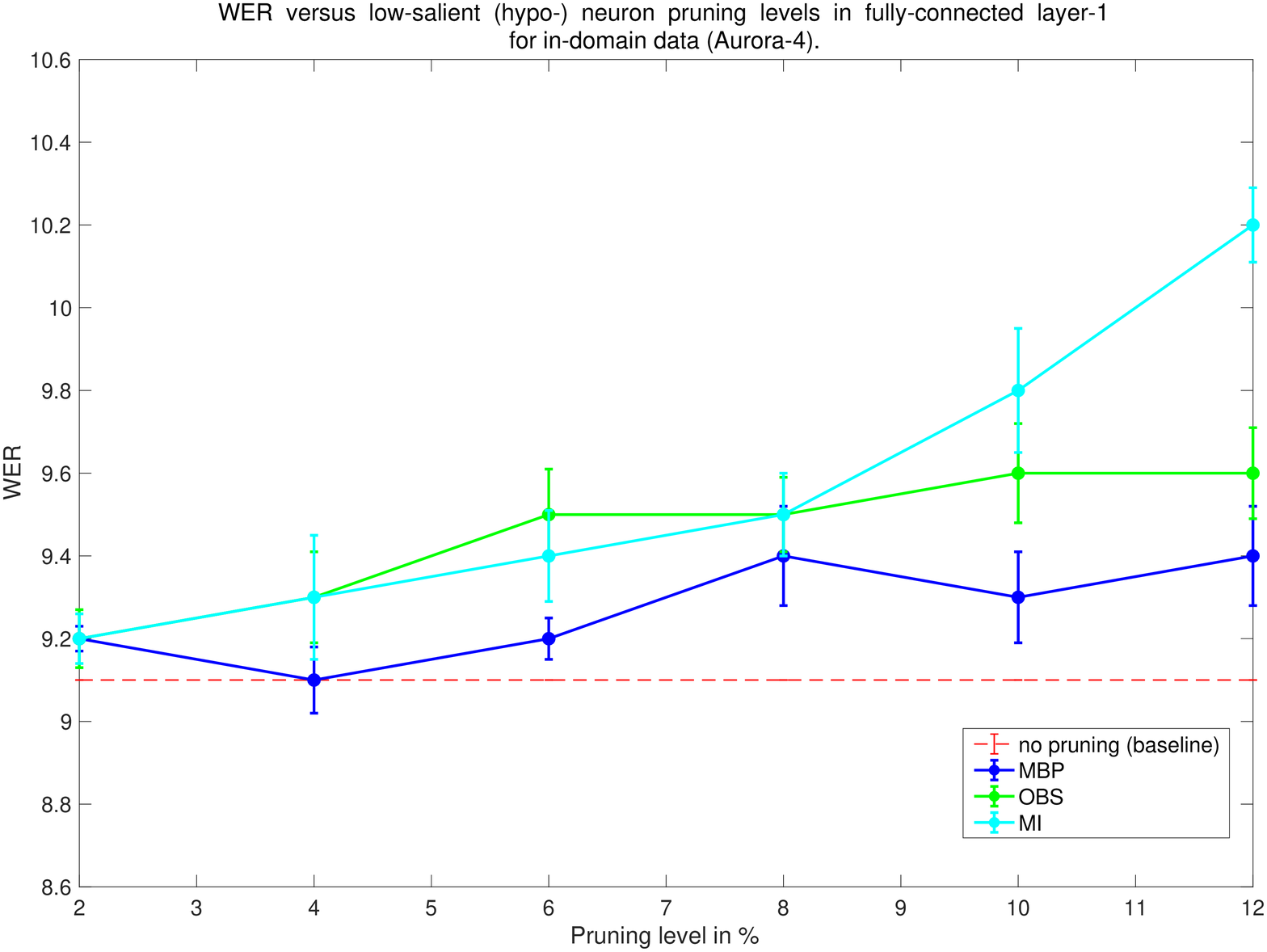}}
\end{minipage}
\caption{WER versus low-salient (hypo-) neuron pruning levels in fully-connected layer-1 for in-domain data (Aurora-4). Red dotted line shows the same for the un-pruned baseline model.}
\label{fig:fig1}
\end{figure}

These are the neurons that are least contributive to information flow within the neural net and have been the candidates for pruning in traditional approaches. We investigated MBP, OBS and MI based pruning of neurons in each fully connected layers of the baseline TFCNN model and analyzed their role on both in-domain and out-of-domain data. The word error rates (WERs) on the Aurora-4 evaluation set, real and simulated dev sets of REVERB14 data is given in Table \ref{tab:table1}, when 2\% pruning is performed only on the 1st layer of the fully connected network. Figure \ref{fig:fig1} shows the outcome of pruning on the $1^{st}$ hidden layer, demonstrating that it did not significantly impact the model’s performance on in-domain data, with up to 6\% relative deterioration at 10\% pruning level; where MBP showed the least deterioration out of the three pruning approaches investigated. 

\begin{table}[th]
%\small
\centering
\caption{WERs (in \%) from the baseline (not pruned) model and the 2\% pruned (fully connected layer 1 only) acoustic model when evaluated on the Aurora-4 and REVERB14 dev (simulated and real condition) set. }
\vspace{5mm}
\label{tab:table1}
  \begin{tabular}{lccccc}
    \toprule
    \multirow{2}{*}{}{Pruning} & Layer & Pruning \% & Aurora-4 &
      \multicolumn{2}{c}{REVERB14} \\
       & {} & {} & {} & {Simulated} & {Real} \\
      \midrule
    None  & - & 0 & 9.1 & 39.3 & 42.4  \\
    MBP   & 1 & 2 & 9.2 (0.0) & 38.5 (0.3) & 41.0 (0.4) \\
    OBS   & 1 & 2 & 9.2 (0.1) & 38.2 (0.1) & 40.1 (0.7) \\
    MI    & 1 & 2 & 9.2 (0.1) & 36.6 (0.3) & 38.9 (0.2) \\
    \bottomrule
  \end{tabular}
\end{table}

\begin{figure}[]
\begin{minipage}[b]{1.0\linewidth}
  \centering
  %\centerline{\includegraphics[width=8.5cm]{draft3_fig1}}
  \centerline{\includegraphics[width=10cm]{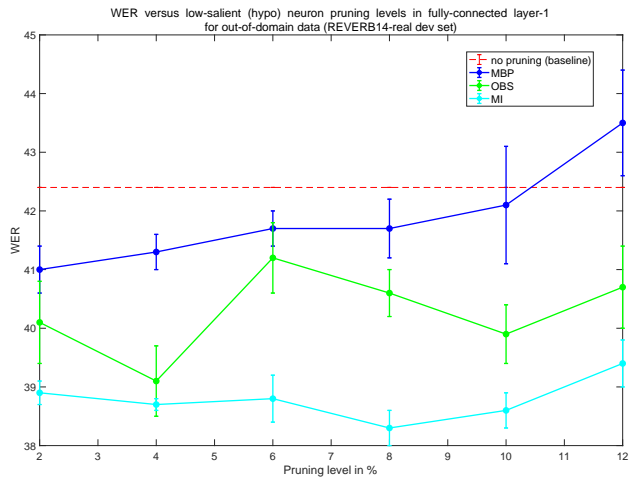}}
\end{minipage}
\caption{WER versus low-salient (hypo) neuron pruning levels in fully-connected layer-1 for out-of-domain data (REVERB14-real dev set). Red dotted line shows the same for the un-pruned baseline model.}
\label{fig:fig2}
\end{figure}

Figure \ref{fig:fig2} shows the effect of $1^{st}$ hidden layer pruning on out-of-domain data. Interestingly, figure \ref{fig:fig2} shows that pruning resulted in performance improvement for out-of-domain data (compared to the unpruned baseline model), with maximum relative reduction of 9.6\% in WER obtained from MI at 8\% pruning level. 

\begin{figure}[]
\begin{minipage}[b]{1.0\linewidth}
  \centering
  %\centerline{\includegraphics[width=8.5cm]{draft3_fig1}}
  \centerline{\includegraphics[width=10cm]{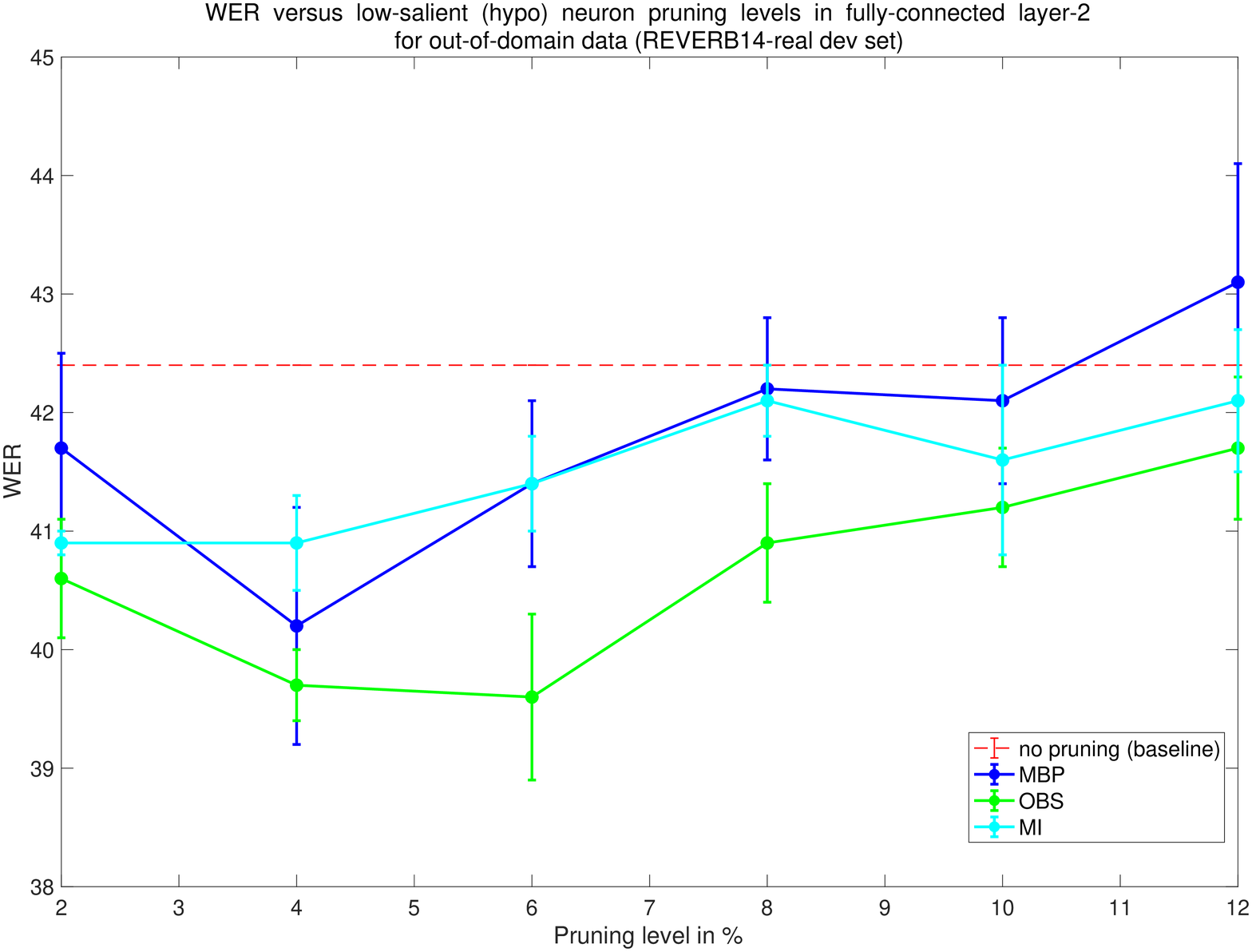}}
\end{minipage}
\caption{WER versus low-salient (hypo) neuron pruning levels in fully-connected layer-2 for out-of-domain data (REVERB14-real dev set). Red dotted line shows the same for the un-pruned baseline model.}
\label{fig:fig3}
\end{figure}

\begin{figure}[]
\begin{minipage}[b]{1.0\linewidth}
  \centering
  %\centerline{\includegraphics[width=8.5cm]{draft3_fig1}}
  \centerline{\includegraphics[width=10cm]{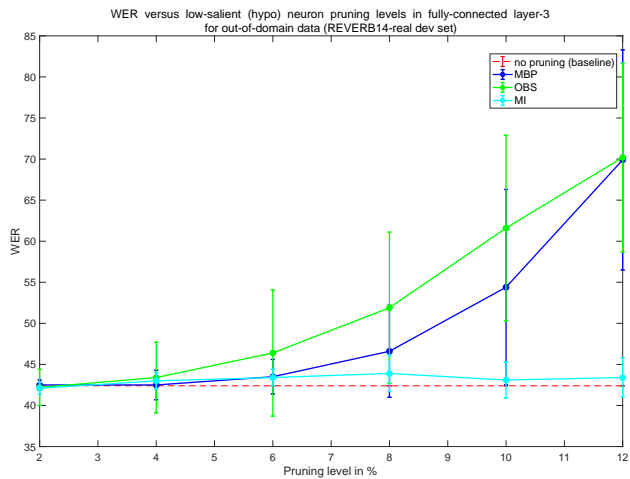}}
\end{minipage}
\caption{WER versus low-salient (hypo) neuron pruning levels in fully-connected layer-3 for out-of-domain data (REVERB14-real dev set). Red dotted line shows the same for the un-pruned baseline model.}
\label{fig:fig4}
\end{figure}

Note that figures \ref{fig:fig1} and \ref{fig:fig2} correspond to pruning only the 1st fully-connected hidden layer. Figures \ref{fig:fig3} and \ref{fig:fig4} show the effect on the performance after pruning the low-salient neurons from the $2^{nd}$ and the $3^{rd}$ fully-connected hidden layer. The performance on in-domain data was found to be similar (as shown in figure \ref{fig:fig1}) for the first two fully connected layers, for up to 12\% pruning. The $3^{rd}$ hidden layer (the prefinal layer) was found to be quite sensitive to pruning (as observed from figure \ref{fig:fig4}), where pruning beyond 6\% neurons resulted in more than 10\% relative increased in WER for in-domain data.

Several interesting observations can be made from figures \ref{fig:fig2}, \ref{fig:fig3} and \ref{fig:fig4}. While MI always performed the best from pruning the 1st and $3^{rd}$ hidden layers, OBS performed the best for pruning the $2^{nd}$ hidden layer. Note that the $3^{rd}$ layer was the pre-final layer, hence pruning beyond 8\% was found to adversely affect the output layer decisions. 

Note that MI uses temporal correlations, and as speech is a quasi-stationary signal, the acoustic features across time have temporal dependencies. The output targets (in this case the senone labels) also have temporal dependencies. The above fact may justify why the performance from MI-based pruning was better across the different layers of the network, compared to the other two approaches. From the above figures the following observations can be made:
\leavevmode
\newline (I) Pruning on first and second hidden layer tend to increase the generalization capacity of the network, which may indicate that there is more redundancy in the hidden layers than in the pre-final layer.
\leavevmode
\newline (II) The pre-final layer seems to be quite sensitive to pruning, where pruning beyond 6\% may significantly reduce the performance of the model. Interestingly, for this case MI-based pruning did not show any performance degradation.
\leavevmode
\newline (III) Removal of low-salient (hypo-neurons) from the network results in improved performance on out-of-domain data-set, while maintaining the performance on in-domain data. Indicating that the model’s generalization capacity can be improved by pruning of the hypo neurons.

\subsection{Pruning of hi-salient (hyper) neurons}
\label{subsection2}

We investigated the role of the high salient (or hyper active) neurons, by layer-wise pruning of such neurons and observing model performance for both in-domain and out-of-domain data. We explored all the three pruning techniques in a layer-wise fashion like the steps in \ref{subsection2}.

\begin{figure}[]
\begin{minipage}[b]{1.0\linewidth}
  \centering
  %\centerline{\includegraphics[width=8.5cm]{draft3_fig1}}
  \centerline{\includegraphics[width=10cm]{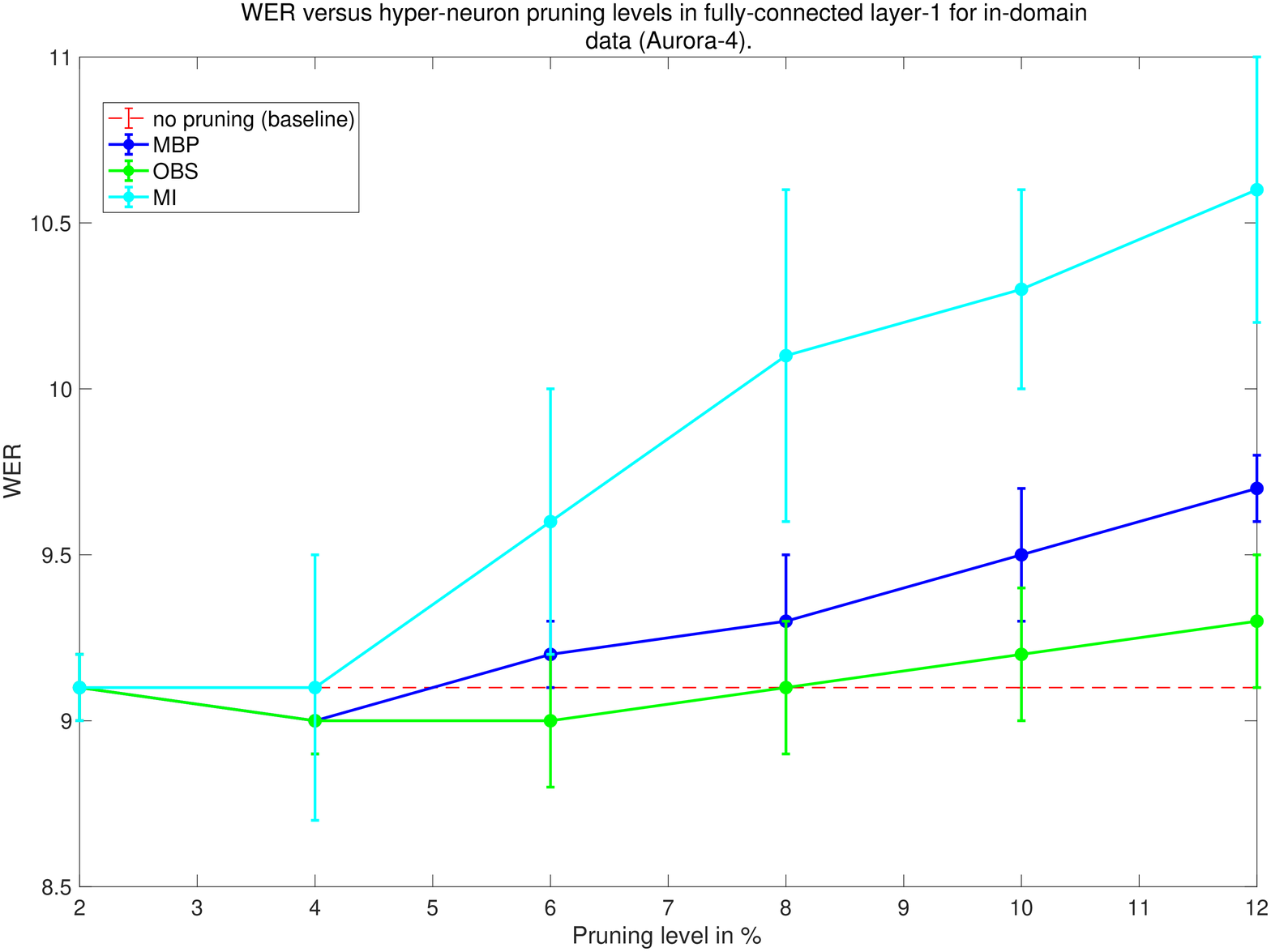}}
\end{minipage}
\caption{WER versus hyper-neuron pruning levels in fully-connected layer-1 for in-domain data (Aurora-4). Red dotted line shows the same for the un-pruned baseline model.}
\label{fig:fig5}
\end{figure}

\begin{figure}[]
\begin{minipage}[b]{1.0\linewidth}
  \centering
  %\centerline{\includegraphics[width=8.5cm]{draft3_fig1}}
  \centerline{\includegraphics[width=10cm]{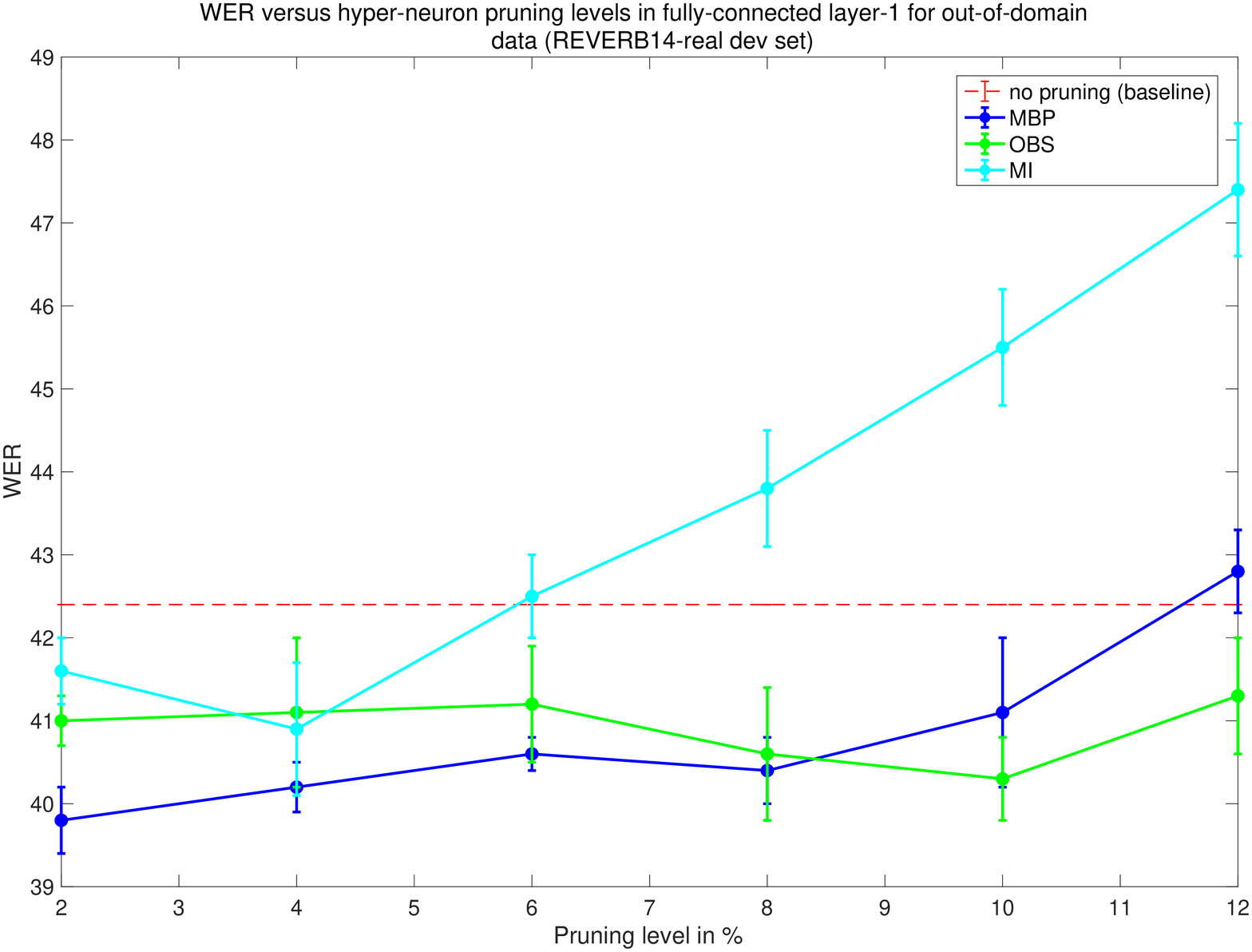}}
\end{minipage}
\caption{WER versus hyper-neuron pruning levels in fully-connected layer-1 for out-of-domain data (REVERB14-real dev set). Red dotted line shows the same for the un-pruned baseline model.}
\label{fig:fig6}
\end{figure}

\begin{figure}[]
\begin{minipage}[b]{1.0\linewidth}
  \centering
  \centerline{\includegraphics[width=10cm]{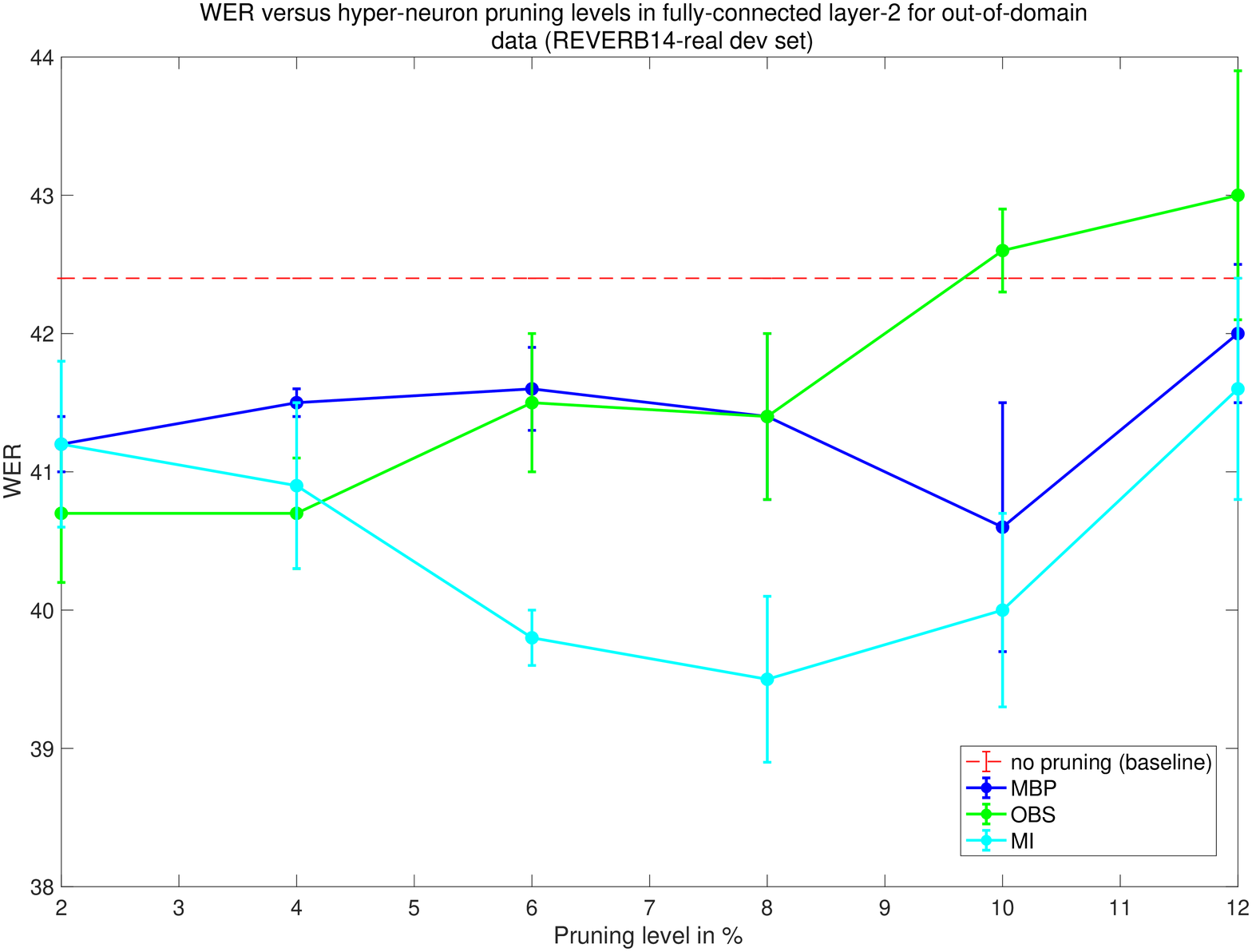}}
\end{minipage}
\caption{WER versus hyper-neuron pruning levels in fully-connected layer-2 for out-of-domain data (REVERB14-real dev set). Red dotted line shows the same for the un-pruned baseline model.}
\label{fig:fig7}
\end{figure}

\begin{figure}[]
\begin{minipage}[b]{1.0\linewidth}
  \centering
  \centerline{\includegraphics[width=10cm]{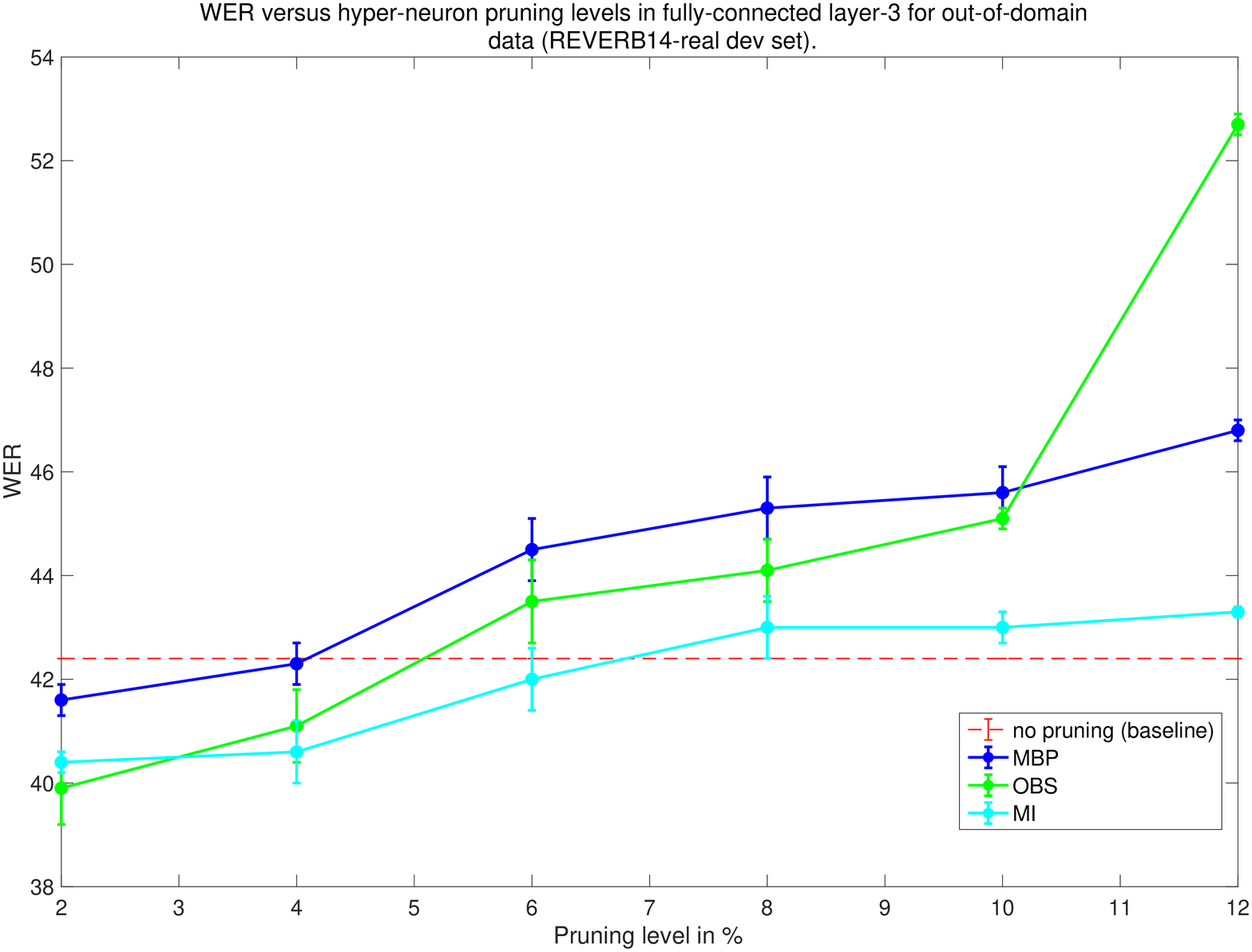}}
\end{minipage}
\caption{WER versus hyper-neuron pruning levels in fully-connected layer-3 for out-of-domain data (REVERB14-real dev set). Red dotted line shows the same for the un-pruned baseline model.}
\label{fig:fig8}
\end{figure}

The WERs on the Aurora-4 evaluation set, and the real dev set of REVERB14 data are shown in figures \ref{fig:fig5} to \ref{fig:fig8}, when pruning is performed from 2\% to 12\% in steps of 2\% for the $1^{st}$, $2^{nd}$ and the $3^{rd}$ fully connected hidden layers.

Figures \ref{fig:fig5} and \ref{fig:fig6} show that MI is the most sensitive amongst the three for hyper-neuron pruning. The model performance reduces for MI after 4\% pruning of hyper-neuron. Whereas for OBS and MBP, up to 10\% pruning of hyper-neuron was possible, which gave better performance on the out-of-domain data, where WER reduction as high as 6.8\% relative to the baseline was observed.

Figure \ref{fig:fig7} shows that for $2^{nd}$ hidden layer, all three pruning techniques resulted in WER reductions, however figure \ref{fig:fig8} shows that the same pruning strategies on the pre-final layer resulted in significant deterioration of performance beyond a pruning threshold of 4\% hyper neurons. `

From figures \ref{fig:fig5} to \ref{fig:fig8} the following observations can be made:
\leavevmode
\newline (I) Pruning of high-salient or hyper neurons can retain baseline performance for in-domain data and can improve performance for out-of-domain data. Hyper neurons are the ones that are more likely to fire given an input feature, hence they may not be contributing to the discriminative power of the network, and thus can be removed without a significant loss in performance. We observed that the removal of up to 4\% of the top hyper neurons can retain in-domain performance while improving the generalization capacity of the network to out-of-domain data. To the best of our knowledge, this is the first claim where hyper active neuron pruning was explored and was found to be useful.
\leavevmode
\newline (II) Salient neurons defined by MI based approach are found to be more essential to the neural net’s performance, than from the other two techniques. Removal of MI-based salient neurons was found to result in significant deterioration of the network’s performance on in-domain data, as compared to OBS and MBP, as observed from figure \ref{fig:fig5}.
\leavevmode
\newline (III) Removal of salient-neurons from the pre-final layer was found to have deleterious effect, where pruning beyond 4\% of top salient features, resulted in performance degradation across all three pruning methods for both in-domain and out-of-domain data.

\subsection{Pruning medium-salient neurons}
\label{subsection3}

Finally, we explored pruning of the medium salient neurons that are neither the lowest 12\% salient neurons nor the top 12\% salient neurons. Exploring layer-wise pruning of mid-salient neurons resulted in catastrophic performance failure (with WER reaching as high as 100\%) for both in-domain and out-of-domain data, indicating that mid-salient neurons are crucial for a neural net’s performance.

While pruning the mid-salient neurons we investigated pruning of the neurons in steps of 10\%, from 5\% pruning to pruning of all mid-salient neurons; and we observed the baseline WER to range from 16.5\% to 100\%. The results indicated that with 5\% pruning of mid-salient neurons the baseline performance deteriorates by more than 80\%.

\subsection{Pruning of both hyper- and hypo- neurons}

Given the observations in sections \ref{subsection1}, \ref{subsection2} and \ref{subsection3}, we investigated layer-wise pruning of both hyper- and hypo- neurons and pruning across multiple layers. Based on findings from sections \ref{subsection1} and \ref{subsection2}, we investigated 8\% pruning of the hypo-neurons and 4\% pruning of the hyper-neurons on layer 1 (that resulted in an overall pruning of 4\% across all the three fully connected layers), and the results are shown in Table \ref{tab:table2}. Table \ref{tab:table2}, shows that pruning both the hyper and hypo neurons in fully connected layer 1, resulted in retaining in-domain performance, while improving the performance on out-of-domain data-set, indicating that such pruning strategy helped to improve the model’s generalization capacity to unseen data conditions. Table \ref{tab:table2} shows that a 4\% pruning of hyper and hypo neurons (based on MI) resulted in WER reduction relative to the baseline of 8.7\% and 7.5\% for simulated and real Reverb14 dev data sets respectively.

\begin{table}[]
%\small
\centering
\caption{WERs from the baseline (not pruned) model and the $8\%$ hypo-neuron and $4\%$ hyper-neuron pruned (deep layer 1 only $\approx 4\%$ pruning across 3 fully-connected layers) acoustic model when evaluated on the Aurora-4 and REVERB14 dev-set.}
\vspace{5mm}
\label{tab:table2}
  \begin{tabular}{lccccc}
    \toprule
    \multirow{2}{*}{}{Pruning} & Layer & Pruning \% & Aurora-4 &
      \multicolumn{2}{c}{REVERB14} \\
      & {} & {} & {} & {Simulated} & {Real} \\
      \midrule
    None  & - & 0 & 9.1       & 39.3       & 42.4  \\
    MBP   & 1 & 4 & 9.2 (0.1) & 38.1 (0.3) & 41.4 (0.9) \\
    OBS   & 1 & 4 & 9.6 (0.3) & 37.5 (0.5) & 40.4 (0.8) \\
    MI    & 1 & 4 & 9.3 (0.1) & 35.9 (0.2) & 39.2 (0.5) \\
    \bottomrule
  \end{tabular}
\end{table}

Next, we investigated the pruning of hyper- and hypo- neurons across multiple layers. We investigated (a) pruning the three fully connected hidden layers and (b) pruning the first two fully connected layers only (ignoring the prefinal layer, as pruning of that layer was not found to be useful from figures \ref{fig:fig4} and \ref{fig:fig8}). Table \ref{tab:table3} shows the results from pruning across multiple layers. Note that for the 1st two fully connected layers 8\% and 4\% hypo and hyper neurons were pruned respectively, and for the pre-final layer 2\% hyper and 2\% hypo neurons were pruned respectively. Table \ref{tab:table3} shows that pruning across all the three layers had more deleterious effect than pruning neurons only from the first two hidden layers, indicating that existing blind pruning techniques may be missing out on potential benefits of selective layer-wise pruning.
Note that overall 8\% pruning was performed for the experimental results shown in table \ref{tab:table3} compared to 4\% overall pruning for the results shown in table \ref{tab:table2}, which justifies the slightly worse out-of-domain performance in table \ref{tab:table3}. 

Table \ref{tab:table3} shows that pruning over first two fully-connected hidden layers gave much better performance compared to pruning across all the fully connected layers or pruning only the first hidden layer. A relative reduction in WER (for the MI based pruning) was found to be 6.7\% and 3.3\%, for the simulated and real sets of out-of-domain (REVERB14 dev) data-set

\begin{table}[th]
%\small
\centering
\caption{WERs from the baseline (not pruned) model and multi-layer pruned (pruning percentage computed over the 3 hidden layers) acoustic model when evaluated on the Aurora-4 and REVERB 2014 dev-set. Percent pruning computed for the entire network.}
\vspace{5mm}
\label{tab:table3}
  \begin{tabular}{lccccc}
    \toprule
    \multirow{2}{*}{}{Pruning} & Layer & Pruning \% & Aurora-4 &
      \multicolumn{2}{c}{REVERB14} \\
      & {} & {} & {} & {Simulated} & {Real} \\
      \midrule
    None  & -     & 0 & 9.1        & 39.3       & 42.4  \\
    MBP   & 1     & 8 & 9.1  (0.2) & 39.1 (0.1) & 43.1 (0.7) \\
    OBS   & 1     & 8 & 9.7  (0.3) & 39.3 (0.7) & 42.1 (0.2) \\
    MI    & 1     & 8 & 9.4  (0.1) & 36.9 (0.3) & 41.2 (0.4) \\
    MBP   & 1-2   & 8 & 9.7  (0.3) & 39.6 (0.3) & 43.7 (0.8) \\
    OBS   & 1-2   & 8 & 9.9  (0.5) & 37.0 (0.5) & 41.7 (0.9) \\
    MI    & 1-2   & 8 & 9.7  (0.3) & 36.7 (0.3) & 41.0 (0.5) \\
    MBP   & 1-2-3 & 9 & 10.4 (0.2) & 41.2 (0.4) & 44.2 (0.8) \\
    OBS   & 1-2-3 & 9 & 10.5 (0.4) & 39.0 (0.5) & 44.3 (0.9) \\
    MI    & 1-2-3 & 9 & 9.9  (0.5) & 37.8 (0.3) & 42.6 (0.6) \\
    \bottomrule
  \end{tabular}
\end{table}

\begin{table}[th]
%\small
\centering
\caption{WERs from the baseline (not pruned) model and multi-layer pruned (pruning percentage computed over the 3 hidden layers) acoustic model when evaluated on the REVERB 2014 test-set. Percent pruning computed for the entire network.}
\vspace{5mm}
\label{tab:table4}
  \begin{tabular}{lcccc}
    \toprule
    \multirow{2}{*}{}{Pruning}& Layer & Pruning \% &
      \multicolumn{2}{c}{REVERB14} \\
      & {} & {} & {Simulated} & {Real} \\
      \midrule
    None  & -     & 0 & 37.8 & 46.9  \\
    MBP   & 1     & 8 & 37.2 & 47.5 \\
    OBS   & 1     & 8 & 36.6 & 46.8 \\
    MI    & 1     & 8 & 35.5 & 46.7 \\
    MBP   & 1-2   & 8 & 36.0 & 44.9 \\
    OBS   & 1-2   & 8 & 35.1 & 43.8 \\
    MI    & 1-2   & 8 & \textbf{34.0} & \textbf{43.4} \\
    MBP   & 1-2-3 & 9 & 37.3 & 46.9 \\
    OBS   & 1-2-3 & 9 & 35.1 & 44.6 \\
    MI    & 1-2-3 & 9 & 34.4 & 44.2 \\
    \bottomrule
  \end{tabular}
\end{table}

Table \ref{tab:table4} presents recognition results from the out-of-domain (REVERB14) data-set, where observations are similar to that in table \ref{tab:table3}. Pruning across the first two fully connected layers were found to be more useful than pruning across all three layers or the first hidden layer alone. A relative reduction in WER (for the MI based pruning) was found to be 10\% and 7.5\%, for the simulated and real sets of out-of-domain (REVERB14 test) data-set.

Following the observations from Tables \ref{tab:table3} and \ref{tab:table4}, we can state the following –
\leavevmode
\newline (a) Pruning both the hyper and hypo neurons can help to improve the generalization capacity of the network.
\leavevmode
\newline (b) The best results on out-of-domain data-set was observed from pruning the fully connected hidden layers without the pre-final layer. The pre-final layer was found to be quite sensitive to pruning for the given task of speech recognition.

Network pruning relies on the fact that a subset of neurons may not have learned reliable information, or, may have learned redundant information, hence preventing the model to generalize well to unseen data conditions. As a consequence, removing such redundant neurons from the network may not cause a significant loss in performance. Table \ref{tab:table3} shows that an 8\% pruning of hyper and hypo neurons across the first two fully connected hidden layers resulted in a relative WER increase of 6.5\% compared to the unpruned baseline model.

Instead of pruning the redundant neurons, they can be updated during unsupervised model adaptation. Typically, during model adaptation, all neurons in one specific layer, or a subset of layers or the whole model are blindly updated, given an adaptation set; where over-fitting is prevented using a suitable regularization parameter. In this work we investigate if there is any benefit in performing initial adaptation of only the redundant neurons (that are supposed to be pruned) and then fine tuning the whole model. Such an unsupervised adaptation technique will focus on learning the new patterns from out-of-domain data through updating the redundant neurons only. 

\section{Acoustic Model Adaptation}

For acoustic model adaptation, we have used the REVERB14 training data-set as the unsupervised adaptation data. The baseline TFCNN acoustic model was used to generate labels for the unsupervised adaptation set and those labels were used to retrain the network. During adaptation, model parameters were updated with an L2 regularization of 0.001 and an initial learning rate of 0.004, with the learning rate halving at every step and a maximum epoch of 10. 

Table \ref{tab:table5} shows the results from the: 
\leavevmode
\newline 
(a) \textbf{Baseline model}: Unpruned TFCNN acoustic model.
\newline (b) \textbf{Model A}: Baseline model after traditional unsupervised adaptation (blind model update), where all parameters of the network were updated.
\leavevmode
\newline (c) \textbf{Model B}:  Baseline model after selective neuron-update based unsupervised model adaptation, where only the redundant neurons (neurons that can be pruned) in the network were updated.
\leavevmode
\newline (d) \textbf{Model C}:  obtained by fine-tuning of Model B, where the adaptation set is augmented with part of the original Aurora-4 training set ($\approx{50}\%$) and all of the model parameters were updated.
\leavevmode
\newline (e) \textbf{Model D}:  obtained by updating all parameters of the baseline model using a mix of the REVERB14 training set and the original Aurora-4 training set ($\approx{50}\%$).

Evaluation of the above five models were performed on the Aurora-4 test set, REVERB-14 real dev and test sets. For unsupervised adaptation, we have used data-selection using the procedure outlined in \cite{mitra2018interpreting}. Table \ref{tab:table5} shows that the selective neuron-update based unsupervised adaptation (Model B) gave better performance than traditional (blind model update) unsupervised adaptation (Model A). Both cases of selective and blind model update resulted in reduction in WER for the out-of-domain data (simulated and real REVERB-2014 dev sets), however, that came at a cost of increase in WER on in-domain data (Aurora-4), due to the phenomena commonly known as the \textit{catastrophic forgetting}. Table \ref{tab:table5} shows that the blind model adaptation had more pronounced performance degradation on in-domain data, which is 68\% relative to the baseline, as compared to the selective neuron-update based unsupervised adaptation, where performance degradation was 38\% relative to the baseline. Hence, Model B suffered less from \textit{catastrophic forgetting} than model A, which can be attributed to the selective model update. Based on the observation from tables 3 and 4, the selective unsupervised adaptation was performed only on the hyper and hypo neurons of layers 1 and 2, based on the decisions from the MI pruning technique. 

The deterioration in the performance on the in-domain data is expected for the selective neuron-update based model adaptation, as only a part of the network gets updated during unsupervised adaptation. Such a selective parameter update can result in altering some network connections that may have had learned patterns in in-domain training data. Model C overcomes such a shift, through the fine-tuning of the Model B by using a part of in-domain training data in addition to unsupervised adaptation data to update the model parameters. After the fine-tuning, Model C demonstrated in-domain performance almost as-good-as the baseline, which can be observed from the final row of table \ref{tab:table5}. 

Additionally, Model D reflects the case where all the parameters of the baseline model is updated with a mix of REVERB14 training set and the Aurora-4 training set. Results from Model D in table \ref{tab:table5} shows that the performance of Model C and Model D are very close to each other, where Model D has a mildly lower WER for REVERB14 eval sets than the Model C. Model-D has the advantage of having all parameter update given the adaptation set and the original training set compared to the selective model update followed by fine tuning in Model C. However, the interesting observation is that the fine-tuning of the model after selective adaptation resulted in a model (Model C) which performed almost as-good-as the blind-model update using both in-domain (Aurora-4) and out-of-domain (REVERB14) data-sets. However, both models C and D assumes the existence of the original training data, which model B does not, hence table 5 clearly shows that in absence of the original training data the selective model update (Model B) can reduce the effect of \textit{catastrophic forgetting} as typically observed in blind model updates (model A). In presence of the original training data both models (selectively trained or not) seems to converge to similar performance.

\begin{table}[th]
%\small
\centering
\caption{WERs from the baseline (not pruned) model, adapted baseline models: Models A, B, C and D (as outlined above), when evaluated on the Aurora-4 and REVERB14 dev set.}
\vspace{5mm}
\label{tab:table5}
  \begin{tabular}{lcccc}
    \toprule
    \multirow{2}{*}{}{Models} & Layer update & Aurora-4 &
      \multicolumn{2}{c}{REVERB14} \\
      & {} & {} & {Simulated} & {Real} \\
      \midrule
    Baseline  & 0    & 9.1  & 39.3 & 42.4  \\
    Model-A   & All  & 15.3 & 24.8 & 33.2 \\
    Model-B   & 1-2  & 12.6 & 23.9 & 32.8 \\
    Model-C   & 1-2  & 9.5  & 22.2 & 32.2 \\
    Model-D   & All  & 9.5  & 22.0 & 31.5 \\

    \bottomrule
  \end{tabular}
\end{table}

Table \ref{tab:table6} shows the results from the baseline model, models A, B, C and D for the REVERB14 test set. Similar to table \ref{tab:table5}, both models A and B show lower WERs for the simulated and real test sets compared to the baseline model. Model B shows better performance than the model A, indicating the effectiveness of the selective-neuron update in unsupervised adaptation. The fine-tuning step resulted in even better performance as observed from the WER obtained from Model C in table \ref{tab:table6}. Fine-tuning lowered the WER by 7.1\% and 0.5\% for the simulated and real test sets respectively, relative to Model B. Selective neuron-update followed by fine-tuning based unsupervised adaptation (model C) resulted in similar performance as the blind model update (Model D) using both REVER14 and Aurora-4 adaptation sets, bolstering the observation from table \ref{tab:table5} showing that the fine-tuning step helped the model to converge to blind model update, where both in-domain and out-of-domain data-sets were used for adaptation. However, in case of the absence of in-domain data for adaptation, the Model B clearly demonstrates that it is relatively resilient to \textit{catastrophic forgetting} as compared to the blind model update.

\begin{table}[th]
%\small
\centering
\caption{WERs from the baseline (not pruned) model, adapted baseline models: Models A, B, C and D (as outlined above), when evaluated on the REVERB 2014 test set.}
\vspace{5mm}
\label{tab:table6}
  \begin{tabular}{lcccc}
    \toprule
    \multirow{2}{*}{}{Models} & Layer update &
      \multicolumn{2}{c}{REVERB14} \\
      & {} & {Simulated} & {Real} \\
      \midrule
    Baseline  & 0    & 37.8 & 46.9  \\
    Model-A   & All  & 22.9 & 37.1 \\
    Model-B   & 1-2  & 22.4 & 35.9 \\
    Model-C   & 1-2  & 20.8 & 35.7 \\
    Model-D   & All  & 20.6 & 34.7 \\

    \bottomrule
  \end{tabular}
\end{table}

\section{Observations}

The goal of this study has been to understand the following: 
\leavevmode
\newline (a) How does neural network pruning help to improve the model’s generalization capacity?
\leavevmode
\newline (b) What are the effects of network pruning in each individual layer of a fully connected network.
\leavevmode
\newline (c) How does the model’s generalization capacity to unseen data vary when we prune the low-salient neurons (that is neurons that are barely active or hypo neurons) as opposed to the high salient neurons (that is neurons that are mostly active or hyper neurons), and if there is any trade-off between choosing one over the other during the network pruning step.
\leavevmode
\newline (d) if there is any benefit of adapting the to-be-pruned neurons during unsupervised adaptation?

Based on the results obtained from our studies, we can make the following observations:
\leavevmode
\newline (i) We observed a significant improvement in an acoustic model’s generalization capacity to out-of-domain data, when pruning was performed at different layers of the fully connected network. We found that pruning of up to 6\% of the neurons can be performed which resulted in retaining in-domain performance, while improving performance on out-of-domain data. Overall an average of 4.5\% relative reduction in WER was observed for out-of-domain data, while no significant change in WER was observed for the in-domain data, from pruning 6\% of the low salient neurons from each of the three layers of the fully connected layer using MI based pruning. 
\leavevmode
\newline (ii) We observed that each individual layer behaved differently to pruning for the given task of acoustic modeling.  Pruning of the first two fully connected layer resulted in significant reduction of WER (with over 97\% confidence), however pruning the pre-final fully connected layer did not result in such significant reduction in WER. The justification of such an observation may be the fact that more redundancy may prevail in the middle layers than the pre-final layer, hence pruning the latter had more deleterious effect on the overall performance of the model.
\leavevmode
\newline (iii) We observed that pruning of low-salient neurons was quite effective in improving the model’s generalization capacity to out-of-domain dataset without sacrificing significant performance degradation on in-domain data. Additionally, we observed that hyper-neurons or overly-active neurons can also be pruned, with similar results. 

Traditionally, the literature has focused on only the pruning of low-salient neurons and this work shows that hyper active neurons can also be pruned from the model without significantly impairing the model’s performance. Experimental results from the pruned acoustic models in our studies demonstrated substantial improvement in recognition results on out-of-domain data as compared to the unpruned model.
\leavevmode
\newline (iv) Finally, we investigated if the to-be-pruned neurons can be selectively updated during unsupervised model adaptation, instead of pruning them out. Results indicated that updating such neurons during unsupervised adaptation results in well regularized networks, that can learn additional information from the out-of-domain adaptation data without significant drop in performance on the in-domain data. Tables 6 shows that the selective neuron update-based model adaptation (model B) resulted in 40.7\% and 23.5\% relative reduction in out-of-domain WER compared to the baseline for simulated and real test sets of REVERB 2014, respectively. The blind model adaptation (model A) provided a 39.4\% and 20.9\% relative reduction in out-of-domain WER compared the baseline for simulated and real test sets of REVERB 2014, respectively. The deterioration in in-domain WER was 38.5\% in case of the selective neuron update based unsupervised adaptation (model B) as opposed to 68.1\% from the blind model update (model A). The fine-tuning stage (updating all the model parameters) after the selective model update resulted in 22.3\% relative reduction in WER for the out-of-domain data with a relative increase in WER of only 4\% for the in-domain data. The improvement in ASR performance from the two-stage unsupervised adaptation was statistically significant when compared to the more traditional blind unsupervised adaptation technique.

\section{Conclusion}

In this work, we investigated neural net pruning, to better understand which neurons can be pruned to improve the generalization capacity of a network to out-of-domain data. We observed that pruning can help in generalization and found that pruning at different layers may have different impacts, where aggressive pruning toward the final layers may not be desirable. We observed that the fully connected layers have broadly three groups of neurons: hypo-neurons, hyper-neurons, and the informative neurons (which are in between the hyper and hypo neurons w.r.t their saliency). Literature primarily focused on pruning off the hypo neurons, but we observed that pruning both hypo and hyper neurons can not only help in reducing the model size, but also improve the model’s generalization capacity. 

Finally, we investigated whether instead of pruning the hyper and hypo neurons, we can selectively update those neurons during unsupervised model adaptation and found such strategy to be more successful compared to blind model update. We observed that selective model update gave better performance on out-of-domain data and diverged less from the in-domain data, compared to the blind model update approach. Finally, a fine-tuning adaptation step using a combination of in-domain and out-of-domain data was found to provide in-domain performance almost as-good-as the in-domain baseline model and a significant relative reduction in out-of-domain WER of more than 22\% (for the real test set of REVERB 2014) compared to the baseline model. 
The findings from this study indicates that during unsupervised adaptation, it may be beneficial to first update the neurons that are deemed as prunable by standard pruning techniques and then perform fine-tuning by updating all the model parameters.

In this work we have focused only on pruning the fully-connected layers, leaving the convolutional layers untouched. From our preliminary investigation we did not find pruning the convolutional layers as-promising-as pruning the fully-connected layer. In future studies, we plan to investigate more on pruning of the convolutional layers,
to see if we can find more room for improvement after the pruning of the fully connected layers.

\section{Acknowledgment}

The authors would like to acknowledge Dimitra Vergyri, the reviewers and the Editor-in-chief of DSP for their helpful thoughts, comments and suggestions.

\bibliography{mybibfile}

\end{document}